%% file: paper.tex
\documentclass[conference]{IEEEtran}
\IEEEoverridecommandlockouts

\usepackage{amsfonts}
\usepackage{url,cite}
\usepackage{graphicx}
\usepackage{booktabs,multirow}
\usepackage{xcolor}

\def\BibTeX{{\rm B\kern-.05em{\sc i\kern-.025em b}\kern-.08em
    T\kern-.1667em\lower.7ex\hbox{E}\kern-.125emX}}
\begin{document}

\title{Improving Pre-Trained Weights Through Meta-Heuristics Fine-Tuning\thanks{The authors are grateful to S\~ao Paulo Research Foundation (FAPESP) grants \#2013/07375-0, \#2014/12236-1, \#2019/07665-4, \#2019/02205-5, and \#2020/12101-0, and to the Brazilian National Council for Research and Development (CNPq) \#307066/2017-7 and \#427968/2018-6.}}

\author{
\IEEEauthorblockN{Gustavo H. de Rosa, Mateus Roder, Jo\~ao Paulo Papa}
\IEEEauthorblockA{Department of Computing \\
S\~ao Paulo State University\\
Bauru - SP, Brazil \\
\{gustavo.rosa, mateus.roder, joao.papa\}@unesp.br}
\and
\IEEEauthorblockN{Claudio F. G. dos Santos}
\IEEEauthorblockA{Department of Computing \\
S\~ao Carlos Federal University\\
S\~ao Carlos - SP, Brazil \\
claudio.santos@ufscar.br}
}

\maketitle

\begin{abstract}
Machine Learning algorithms have been extensively researched throughout the last decade, leading to unprecedented advances in a broad range of applications, such as image classification and reconstruction, object recognition, and text categorization. Nonetheless, most Machine Learning algorithms are trained via derivative-based optimizers, such as the Stochastic Gradient Descent, leading to possible local optimum entrapments and inhibiting them from achieving proper performances. A bio-inspired alternative to traditional optimization techniques, denoted as meta-heuristic, has received significant attention due to its simplicity and ability to avoid local optimums imprisonment. In this work, we propose to use meta-heuristic techniques to fine-tune pre-trained weights, exploring additional regions of the search space, and improving their effectiveness. The experimental evaluation comprises two classification tasks (image and text) and is assessed under four literature datasets. Experimental results show nature-inspired algorithms' capacity in exploring the neighborhood of pre-trained weights, achieving superior results than their counterpart pre-trained architectures. Additionally, a thorough analysis of distinct architectures, such as Multi-Layer Perceptron and Recurrent Neural Networks, attempts to visualize and provide more precise insights into the most critical weights to be fine-tuned in the learning process.
\end{abstract}


\input{sections/introduction.tex}
\input{sections/ml.tex}
\input{sections/opt.tex}
\input{sections/methodology.tex}
\input{sections/experiments.tex}
\input{sections/conclusion.tex}

\bibliographystyle{IEEEtran}
\bibliography{references}

\end{document}

%% file: sections/introduction.tex
\section{Introduction}
\label{s.introduction}

Intelligence-based systems brought better insights into decision-making tasks and withdrew part of the humans' burden in recurring tasks, where most of these advances have arisen from research fostered by Artificial Intelligence (AI)~\cite{Lu:19} and Machine Learning (ML)~\cite{Boutaba:18}. They have been incorporated in a wide range of applications, such as autonomous driving~\cite{Yurtsever:20}, text classification~\cite{Kowsari:19}, image and object recognition~\cite{Druzhkov:16}, and medical analysis~\cite{Litjens:17}, among others.

The increasing demand for more complex tasks and the ability to solve unprecedented problems strengthened an ML sub-area, denoted as Deep Learning (DL)~\cite{Pouyanfar:18}. DL algorithms are known for employing deep neural networks and millions of parameters to model the intrinsic nature of the human brain~\cite{Kriegeskorte:19}, i.e., learn how humans can process information through their visual system and how they can communicate between themselves. Nevertheless, such learning is conditioned to the training data and the model's parameters and often does not reproduce the real-world environment, leading to undesired behavior, known as underfitting/overfitting~\cite{Belkin:18}.

Even though proper training usually accompanies underfitting/overfitting, it is common to perceive a poor performance when the model is collated with unseen data. This discrepancy lies in the fact that the model ``memorized" the training data instead of learning its patterns, thus not reproducing the desired outputs when applied to slightly-different data (test data). The best approach to overcome this problem would be to employ combinations of all possible parameters and verify whether they are suitable or not when applied to the testing data. Nevertheless, such an approach is unfeasible regarding DL architectures due to their vast number of parameters and exponential complexity~\cite{Xiong:11}.

On the other hand, a more feasible approach stands for optimization procedures, where parameters are optimized according to an objective function instead of being joined in all possible combinations. A recent technique, denoted as meta-heuristic, has attracted considerable attention in the last years, mainly due to its simple heuristics and ability to optimize non-differentiable functions. For instance, Rosa et al.~\cite{Rosa:15} used the Harmony Search algorithm for fine-tuning Convolutional Neural Networks (CNN) hyperparameters, achieving improved results over the benchmark architectures. At the same time, Rodrigues et al.~\cite{Rodrigues:17} explored single- and multi-objective meta-heuristic optimization in Machine Learning problems, such as feature extraction and selection, hyperparameter tuning, and unsupervised learning. Furthermore, Wang et al.~\cite{Wang:19} presented a fast-ranking version of the Particle Swarm Optimization algorithm to fine-tune CNN hyperparameters and remove the fitness function evaluation cost.

Nevertheless, most of the literature works focus on only optimizing the model's hyperparameters (learning rate, number of units, momentum, weight decay, dropout)~\cite{Bergstra:15,Yao:17,Rosa:19,Han:20} instead of optimizing its parameters (layers' weights and biases)~\cite{Nawi:19}. Usually, parameters are optimized during the learning procedure through gradient-based approaches, such as the Stochastic Gradient Descent, yet they might benefit from the meta-heuristic techniques' exploration and exploitation capabilities.

This work proposes an additional fine-tuning after the model's training, aiming to explore unknown search space regions that gradient-based optimizers could not find. Such an approach is conducted by exploring weights under pre-defined bounds and evaluating them according to an objective function (accuracy over the validation set). Therefore, the main contributions of this work are three-fold: (i) to introduce meta-heuristic optimization directly to the model's parameters, (ii) to provide insightful analysis of whether gradient-based optimizers achieved feasible regions or not, and (iii) to fill the lack of research regarding meta-heuristic optimization applied to Machine Learning algorithms.

The remainder of this paper is organized as follows. Section~\ref{s.ml} presents a theoretical background regarding the employed ML architectures, e.g., Multi-Layer Perceptrons and Recurrent Neural Networks. Section~\ref{s.opt} introduces a brief explanation about meta-heuristic optimization, as well as the Genetic Algorithm and Particle Swarm Optimization. Section~\ref{s.methodology} presents the mathematical formulation of the proposed approach, its complexity analysis, the employed datasets, and the experimental setup. Finally, Section~\ref{s.experiments} discusses the experimental results while Section~\ref{s.conclusion} states the conclusions and future works.

%% file: sections/ml.tex
\section{Machine Learning}
\label{s.ml}

This section introduces brief concepts regarding the Multi-Layer Perceptron and the Long Short-Term Memory.

\subsection{Multi-Layer Perceptron}
\label{ss.mlp}

\begin{figure}[!ht]
	\centering
	\includegraphics[scale=0.4]{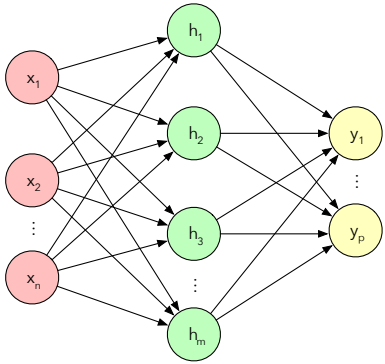}
	\caption{Multi-Layer Perceptron standard architecture.}
	\label{f.mlp}	
\end{figure}

Multi-Layer Perceptron has arisen from the traditional Perceptrons and represents a type of feed-forward artificial neural network. Instead of having a single intermediate layer and a linear function as the Perceptron has, the MLP architecture comprises multiple Perceptrons arranged in hidden layers and followed by non-linear activations, which allows it to distinguish non-linearly separable data. Figure~\ref{f.mlp} illustrates the standard architecture of an MLP.

Commonly, MLP-based networks are employed in supervised learning tasks and trained through the Backpropagation~\cite{Rumelhart:86} algorithm, which calculates the output error and corrects the model's weights according to the derivative of the activation function and part of the error. Modern MLP architectures use more sophisticated activation functions, such as the Rectified Linear Unit (ReLU)~\cite{Nair:10}, instead of only relying on traditional ones, e.g., sigmoid and hyperbolic tangents. Additionally, they have been used as the foundation of several Deep Learning architectures, such as VGG~\cite{Simonyan:14} and Inception~\cite{Szegedy:16} ones.

\subsection{Long Short-Term Memory}
\label{ss.rnn}

\begin{figure}[!ht]
	\centering
	\includegraphics[scale=0.35]{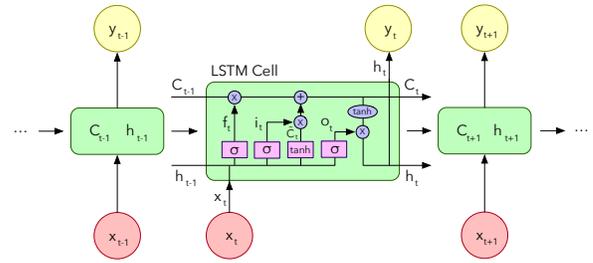}
	\caption{Long Short-Term Memory cell architecture.}
	\label{f.lstm}	
\end{figure}

Hochreiter et al.~\cite{Hochreiter:97} proposed the Long Short-Term Memory networks, which are particular types of Recurrent Neural Networks designed to learn information through long periods. Their main difference when compared to traditional RNNs lies in their hidden layer, which employs a cell (unit) with four gate mechanism that interacts between themselves. Figure~\ref{f.lstm} illustrates such a cell architecture.

An LSTM cell is represented by the following variables: cell state $\mathbf{C}_t \in \mathbb{R}^k$, input neurons $\mathbf{x}_t \in \mathbb{R}^n$, hidden layer $\mathbf{h}_t \in \mathbb{R}^k$ and output neurons $\mathbf{y}_t \in \mathbb{R}^p$. The cell state works as a conveyor, running through the belt and suffering linear combinations. Additionally, LSTM's cell may add or remove self-contained information through its gate mechanisms.

The gates allow or deny the flow of information and usually constitute non-linear neural layers, such as sigmoid activation and point-wise operations. The output of a sigmoid function creates a real number between $0$ and $1$, which describes the amount of information the gate should propagate. Note that $0$ stands for no-information, while $1$ stands for the full-information. Commonly, an LSTM cell is regulated by three gates: input, forget, and output.

%% file: sections/opt.tex
\section{Meta-Heuristic Optimization}
\label{s.opt}

Optimization problems consist of maximizing or minimizing mathematical functions through potential values, while the optimization procedure aims to find those values given a pre-defined domain. Traditional optimization methods~\cite{Bertsekas:99}, such as the combinatorial and iterative methods, such as the Grid-Search, the Newton method, the Quasi-Newton, the Gradient Descent, Interpolation methods, use the evaluation of gradients and Hessians, thus, being only practical when applied to differentiable functions. Additionally, they elevate the computational burden due to calculating first- and second-order derivatives. 

Alternatively, an approach known as meta-heuristic has been applied to solve optimization tasks. Meta-heuristic techniques~\cite{Yang:11} consists of high-level algorithms designed to create and select heuristics capable of producing feasible solutions to the optimization problem. Hence, meta-heuristic optimization is a procedure that connects notions of \emph{exploration}, used to conduct extensive searches throughout the space, and \emph{exploitation}, used to improve potential solutions based on their neighborhoods.

\subsection{Genetic Algorithm}
\label{ss.ga}

Genetic Algorithm is a traditional evolutionary-based algorithm inspired by the process of natural selection. It commonly relies on biological operators, such as selection, mutation, and crossover, and generates feasible solutions for optimization tasks. Each individual is represented by an $n$-dimensional position array $\mathbf{x}$, where each dimension stands for a decision variable, and a fitness value associated with this particular position, i.e., $f(\mathbf{x})$. 

The Genetic Algorithm's main objective is to evolve a population of $m$ individuals in an iterative way, where the so-called biological operators are applied over the population to create a more fit population, e.g., lower fitness values in minimization problems. During each iteration/generation, the population is evaluated, and a set of $p_s \times m$ individuals are stochastically selected from it, where $p_s$ stands for the selection proportion selection and $m$ the number of individuals. 

Furthermore, the selected individuals are divided into pairs to form the ``parents" and bred into offsprings according to a crossover probability, denoted as $p_c$. The offsprings are new individuals who share characteristics inherited from their parents, i.e., they have randomly selected positions from their mother and father. Afterward, the offsprings are mutated according to a mutation probability $p_m$, which occasionally adds a noise value to one of the offsprings' positions.

Finally, the population is re-evaluated, and the iterative process continues until a stop criterion is satisfied, such as an epsilon or a maximum number of generations. Combining the biological operators' explorability and exploitability allows the population to convergence to more appropriate values, thus producing feasible solutions to optimization tasks.

\subsection{Particle Swarm Optimization}
\label{ss.swarm}

Particle Swarm Optimization is a nature-inspired algorithm that designs each agent as a bird that belongs to a swarm and searches for optimal food sources. Each agent is represented by a $(\mathbf{x}, \mathbf{v})$ tuple, where $\mathbf{x}$ stands for its position and $\mathbf{v}$ for its velocity. The initial position $\mathbf{x}$ is described as an $n$-dimensional randomly vector, while the velocity $\mathbf{v}$ is an $n$-dimensional vector of zeros, where each dimension stands for the decision variable. Additionally, the objective corresponds to searching the most likely decision variables, which maximizes or minimizes a target function.

Let $\mathbf{v}_i^t$ be the velocity of an agent $i$ at iteration $t$, belonging to a swarm of size $M$, such that $i \in \{1, 2, \ldots, M\}$. One can update its velocity according to Equation~\ref{e.velocity}, as follows:

\begin{equation}
\label{e.velocity}
\mathbf{v}_i^{t+1} = w \mathbf{v}_i^t + c_1 r_1 (\mathbf{x}_i^{*}-\mathbf{x}_i^t) + c_2 r_2 (\mathbf{g}-\mathbf{x}_i^t),
\end{equation}
where $\mathbf{x}_i^{*}$ stands for the best position obtained by agent $i$, and $\mathbf{g}$ denotes the current best solution. Additionally, $w$, $c_1$, and $c_2$ stand for the inertia weight, the social parameter, and the cognitive ratio, respectively. Finally, $r_1$ and $r_2$ are uniformly distributed random numbers in the range $[0,1]$. 

Furthermore, let $\mathbf{x}_i^t$ be the position of an agent $i$ at iteration $t$. One can update its position according to Equation~\ref{e.position}, as follows:

\begin{equation}
\label{e.position}
\mathbf{x}_i^{t+1} = \mathbf{x}_i^t + \mathbf{v}_i^{t+1}.
\end{equation}

%% file: sections/methodology.tex
\section{Methodology}
\label{s.methodology}

This section presents a brief discussion regarding the proposed approach, its complexity analysis, the employed datasets, and the experimental setup.

\subsection{Proposed Approach}
\label{s.proposed}

The proposed approach aims to pre-train an architecture through its standard pipeline, e.g., stochastic gradient optimization across a training set, followed by a fine-tuning using meta-heuristic optimization across a validation set (post-trained). The idea is to use meta-heuristic techniques to explore the search space better and intensify a promising solution found by the traditional optimization algorithm.

\begin{figure}[!ht]
	\centering
	\includegraphics[scale=0.5]{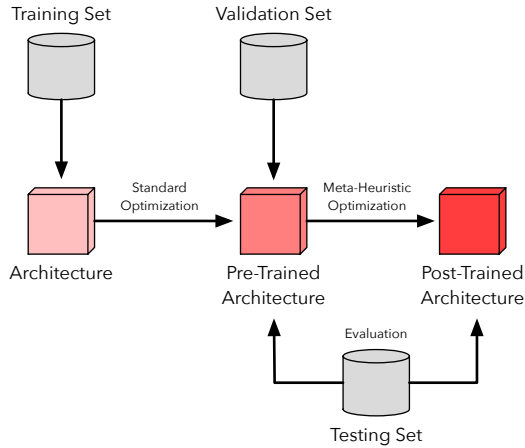}
	\caption{Proposed approach pipeline.}
	\label{f.pipeline}	
\end{figure}

Let $\theta$ be the weights of a pre-trained neural network, where $\theta \in \Re^n$ and $n$ stands for the number of weights. Additionally, let $\Delta$ be a defined value which stands for the search bounds around $\theta$. The initial solutions are randomly sampled from the $[\theta-\Delta, \theta+\Delta$] interval\footnote{Note that although the search interval ($\Delta$) is equal for each variable in $\theta$, they may have a different value.} and feed to the meta-heuristic technique, which will explore the search space and find the most suitable solutions given a fitness function, i.e., accuracy over the validation set. At the end of the optimization procedure, both post-trained and pre-trained networks are evaluated over testing sets and compared. Figure~\ref{f.pipeline} illustrates an overview of the proposed approach pipeline.

\subsection{Complexity Analysis}
\label{ss.complexity}

Let $O(\iota)$ and $O(\zeta)$ be the complexity of training and validating a network for each epoch, respectively. We can observe that in the proposed approach, we opted to pre-train the whole network using $T_n$ iterations, with additional validations for every epoch. Hence, the whole pre-training procedure complexity is depicted by Equation~\ref{e.pre_train_complexity}, as follows:

\begin{equation}
\label{e.pre_train_complexity}
O(\iota) \cdot T_n + O(\zeta) \cdot T_n = O(\iota + \zeta) \cdot T_n.
\end{equation}

The proposed approach intends to provide an additional optimization step after the network's pre-training, where agents will encode the pre-trained weights and biases as their positions, search for better solutions throughout the space and evaluate the fitness function (validation). Let $T_o$ be the number of optimization iterations, $m$ the number of agents, and the whole optimization procedure complexity described by Equation~\ref{e.opt_complexity}, as follows:

\begin{equation}
\label{e.opt_complexity}
O(\zeta) \cdot T_o \cdot m.
\end{equation}

Therefore, summing both Equations~\ref{e.pre_train_complexity} and~\ref{e.opt_complexity} together, it is possible to achieve the proposed approach complexity, as described by Equation~\ref{e.total_complexity}:

\begin{equation}
\label{e.total_complexity}
O(\iota + \zeta) \cdot T_n + O(\zeta) \cdot T_o \cdot m.
\end{equation}

Note that if $O(\zeta) \rightarrow 0$, both approaches will have the same complexity. Unfortunately, this pattern will not often happen, mainly due to the complexity of larger models, such as CNNs and RNNs. However, the advantage is that $O(\zeta)$ is significantly smaller than $O(\iota)$, thus adding only a small burden over the standard pre-training.

\subsection{Datasets}
\label{ss.datasets}

We considered four datasets in the experimental section, being two image- and two text-based:

\begin{itemize}
	\item CIFAR-10~\cite{Krizhevsky:09}: is a subset image database from the ``80 million tiny images" dataset. Composed of $60,000$ 32x32 colour images divided in 10 classes, with $6,000$ images per class. It is divided into five training batches and one test batch, each containing $10,000$ images;
	\item CIFAR-100~\cite{Krizhevsky:09}: is almost like the CIFAR-10 dataset, yet it provides a more challenging problem. Composed of $60,000$ 32x32 colour images divided in 100 classes, with $600$ images per class. It is also split into five training batches and one test batch, each containing $10,000$ images;
	\item IMDB Large Movie Reviews~\cite{Maas:11}: is a dataset for the task of sentiment analysis (binary classification) and is composed of $50,000$ raw-text movie reviews, equally split into training and testing sets;
	\item Stanford Sentiment Treebank (SST)~\cite{Socher:13}: is composed of three-level sentiment analysis (positive, negative, and neutral) of syntactically plausible phrase in thousands of sentences from Rotten Tomatoes movie reviews (more than $200,000$).
\end{itemize}

\subsection{Experimental Setup}
\label{ss.setup}

The proposed approach is evaluated amongst two distinct tasks: image classification (CIFAR-10 and CIFAR-100) and sentiment analysis (SST and IMDB). Regarding the former task, we employed a standard Multi-Layer Perceptron (input, hidden, and output layers), while the latter task uses a Long Short-Term Memory (embedding, hidden, and output layers). Table~\ref{t.params} describes the hyperparameters used in the image- and text-based datasets, respectively.

\begin{table}[!ht]
	\centering
	\setlength{\tabcolsep}{5pt}
	\renewcommand{\arraystretch}{1.5}
	\caption{Hyperparameters configuration used in image classification (MLP) and sentiment analysis (LSTM) tasks.}
	\label{t.params}
	\begin{tabular}{lcccc}
		\toprule
		\textbf{Hyperparameter} & \textbf{CIFAR-10} & \textbf{CIFAR-100} & \textbf{IMDB} & \textbf{SST}
		\\ \midrule
		$n_i$ (input units) & $3,072$ & $3,072$ & $-$ & $-$
		\\
		$n_e$ (embedding units) & $-$ & $-$ & $128$ & $64$
		\\
		$n$ (hidden units) & $1,024$ & $2,048$ & $512$ & $128$
		\\
		$n_o$ (output units) & $10$ & $100$ & $2$ & $3$
		\\
		$e$ (epochs) & $50$ & $50$ & $10$ & $20$
		\\
		$bs$ (batch size) & $100$ & $100$ & $128$ & $8$
		\\
		$\eta$ (learning rate) & $0.0001$ & $0.0001$ & $0.01$ & $0.0001$
		\\ \bottomrule
	\end{tabular}
\end{table}

Regarding the meta-heuristic techniques, we opted to use an evolutionary-based algorithm denoted as Genetic Algorithm and a swarm-based one, known as Particle Swarm Optimization. Both algorithms are available in the Opytimizer~\cite{Rosa:19} package and the paper's source code at GitHub\footnote{The source code is available at \url{https://github.com/gugarosa/mh_fine_tuning}.}. For each meta-heuristic, we employed three search space configurations, as follows:

\begin{itemize}
	\item $\alpha$: $10$ agents optimized over $5$ iterations;
	\item $\beta$: $50$ agents optimized over $25$ iterations;
	\item $\gamma$: $100$ agents optimized over $50$ iterations.
\end{itemize}

Additionally, Table~\ref{t.opt_param} describes the meta-heuristics parameters configuration.

\begin{table}[!ht]
	\centering
	\setlength{\tabcolsep}{5pt}
	\renewcommand{\arraystretch}{1.5}
	\caption{Meta-heuristics parameters configuration.}
	\label{t.opt_param}
	\begin{tabular}{lc}
		\toprule
		\textbf{Meta-heuristic} & \textbf{Parameters} 
		\\ \midrule
		GA & $p_s=0.75 \mid p_c=0.5 \mid p_m=0.25$
		\\
		PSO & $w=0.7 \mid c_1=1.7 \mid c_2=1.7$
		\\ \bottomrule
	\end{tabular}
\end{table}

Finally, to provide a more robust analysis, we conduct $10$ runnings with different seeds (different splits) for each (metaheuristic, dataset) pair, followed by a statistical analysis according to the Wilcoxon signed-rank test~\cite{Wilcoxon:45} with a $0.05$ significance. 

%% file: sections/experiments.tex
\section{Experiments and Discussion}
\label{s.experiments}

This section presents the experimental results concerning the employed datasets and tasks. Furthermore, we present additional discussions regarding the effect on weights' optimization and how the search space's bounds influence the fine-tuning.

\subsection{Overall Discussion}
\label{ss.overall}

Table~\ref{t.image_results} describes the experimental results over the image classification task, which comprehends both CIFAR-10 (top) and CIFAR-100 (bottom) datasets. The most striking point to elucidate is that almost every meta-heuristic model could slightly improve the baseline accuracy, i.e., every meta-heuristic on the CIFAR-10 dataset, as well as $\alpha$-GA-MLP and $\beta$-PSO-MLP on the CIFAR-100 dataset. Additionally, considering both datasets, $\alpha$-GA-MLP obtained the best accuracy and recall metrics amongst the evaluated models (underlined cells), yet every evaluated model was statistically similar according to the Wilcoxon signed-rank test (bolded cells).

\begin{table*}[!ht]
	\centering
	\setlength{\tabcolsep}{10pt}
	\renewcommand{\arraystretch}{1.75}
	\caption{Experimental results ($\%$) over CIFAR-10 (top) and CIFAR-100 (bottom) datasets.}
	\label{t.image_results}
	\begin{tabular}{lcccc}
		\toprule
		\textbf{Model} & \textbf{Accuracy} & \textbf{Precision} & \textbf{Recall} & \textbf{F1-Score}
		\\ \midrule
		MLP & $\mathbf{52.40 \pm 0.62}$ & $\mathbf{52.73 \pm 0.57}$ & $\mathbf{52.40 \pm 0.62}$ & $\mathbf{52.45 \pm 0.60}$
		\\
		$\alpha$-GA-MLP & $\mathbf{\underline{52.54 \pm 0.52}}$ & $\mathbf{\underline{52.76 \pm 0.54}}$ & $\mathbf{\underline{52.54 \pm 0.52}}$ & $\mathbf{\underline{52.54 \pm 0.52}}$
		\\
		$\alpha$-PSO-MLP & $\mathbf{52.51 \pm 0.56}$ & $\mathbf{52.72 \pm 0.57}$ & $\mathbf{52.51 \pm 0.56}$ & $\mathbf{52.50 \pm 0.56}$
		\\
		$\beta$-GA-MLP & $\mathbf{52.49 \pm 0.57}$ & $\mathbf{52.70 \pm 0.57}$ & $\mathbf{52.49 \pm 0.57}$ & $\mathbf{52.48 \pm 0.58}$
		\\
		$\beta$-PSO-MLP & $\mathbf{52.53 \pm 0.56}$ & $\mathbf{52.73 \pm 0.57}$ & $\mathbf{52.53 \pm 0.56}$ & $\mathbf{52.52 \pm 0.56}$
		\\
		$\gamma$-GA-MLP & $\mathbf{52.52 \pm 0.59}$ & $\mathbf{52.73 \pm 0.62}$ & $\mathbf{52.52 \pm 0.59}$ & $\mathbf{52.51 \pm 0.60}$
		\\
		$\gamma$-PSO-MLP & $\mathbf{52.52 \pm 0.59}$ & $\mathbf{52.73 \pm 0.61}$ & $\mathbf{52.52 \pm 0.59}$ & $\mathbf{52.51 \pm 0.60}$
		\\ \midrule \midrule
		MLP & $\mathbf{24.93 \pm 0.30}$ & $\mathbf{25.59 \pm 0.35}$ & $\mathbf{24.93 \pm 0.30}$ & $\mathbf{24.74 \pm 0.29}$
		\\
		$\alpha$-GA-MLP & $\mathbf{\underline{24.96 \pm 0.33}}$ & $\mathbf{25.64 \pm 0.40}$ & $\mathbf{\underline{24.96 \pm 0.33}}$ & $\mathbf{24.77 \pm 0.32}$
		\\
		$\alpha$-PSO-MLP & $\mathbf{24.93 \pm 0.31}$ & $\mathbf{25.59 \pm 0.33}$ & $\mathbf{24.93 \pm 0.31}$ & $\mathbf{24.73 \pm 0.30}$
		\\
		$\beta$-GA-MLP & $\mathbf{24.91 \pm 0.33}$ & $\mathbf{25.63 \pm 0.40}$ & $\mathbf{24.91 \pm 0.33}$ & $\mathbf{24.75 \pm 0.33}$
		\\
		$\beta$-PSO-MLP & $\mathbf{\underline{24.96 \pm 0.34}}$ & $\mathbf{25.61 \pm 0.37}$ & $\mathbf{\underline{24.96 \pm 0.34}}$ & $\mathbf{24.76 \pm 0.32}$
		\\
		$\gamma$-GA-MLP & $\mathbf{24.92 \pm 0.31}$ & $\mathbf{\underline{25.65 \pm 0.38}}$ & $\mathbf{24.92 \pm 0.31}$ & $\mathbf{\underline{24.78 \pm 0.32}}$
		\\
		$\gamma$-PSO-MLP & $\mathbf{24.92 \pm 0.31}$ & $\mathbf{25.52 \pm 0.39}$ & $\mathbf{24.92 \pm 0.31}$ & $\mathbf{24.70 \pm 0.31}$
		\\ \bottomrule
	\end{tabular}
\end{table*}

Table~\ref{t.text_results} describes the experimental results over the sentiment classification task, which comprehends both IMDB (top) and SST (bottom) datasets. The meta-heuristics performance was marginally inferior to the baseline architecture (best results marked by underlined cells) considering all metrics and models. Such behavior is possibly explained by the fact that meta-heuristics optimized the last fully-connected layer, which is not the most descriptive layer in a recurrent network. Even though they did not achieve outstanding results, every model has been statistically similar to the baseline experiment according to the Wilcoxon signed-rank test (bolded cells) and reinforces the meta-heuristics capacity in attempting to search for more reasonable minimum points.

\begin{table*}[!ht]
	\centering
	\setlength{\tabcolsep}{10pt}
	\renewcommand{\arraystretch}{1.75}
	\caption{Experimental results ($\%$) over IMDB (top) and SST (bottom) datasets.}
	\label{t.text_results}
	\begin{tabular}{lcccc}
		\toprule
		\textbf{Model} & \textbf{Accuracy} & \textbf{Precision} & \textbf{Recall} & \textbf{F1-Score}
		\\ \midrule
		LSTM & $\mathbf{\underline{49.21 \pm 2.05}}$ & $\mathbf{\underline{49.17 \pm 1.69}}$ & $\mathbf{\underline{49.21 \pm 1.23}}$ & $\mathbf{\underline{48.54 \pm 1.98}}$
		\\
		$\alpha$-GA-LSTM & $\mathbf{48.98 \pm 1.76}$ & $\mathbf{48.91 \pm 1.75}$ & $\mathbf{48.98 \pm 1.62}$ & $\mathbf{48.14 \pm 1.81}$
		\\
		$\alpha$-PSO-LSTM & $\mathbf{48.97 \pm 1.54}$ & $\mathbf{48.93 \pm 1.60}$ & $\mathbf{48.92 \pm 1.81}$ & $\mathbf{48.10 \pm 1.90}$
		\\
		$\beta$-GA-LSTM & $\mathbf{48.98 \pm 1.93}$ & $\mathbf{48.91 \pm 1.90}$ & $\mathbf{48.98 \pm 1.97}$ & $\mathbf{48.14 \pm 1.77}$
		\\
		$\beta$-PSO-LSTM & $\mathbf{48.98 \pm 1.88}$ & $\mathbf{48.90 \pm 1.78}$ & $\mathbf{48.97 \pm 1.99}$ & $\mathbf{48.16 \pm 1.73}$
		\\
		$\gamma$-GA-LSTM & $\mathbf{49.03 \pm 1.88}$ & $\mathbf{48.96 \pm 1.77}$ & $\mathbf{49.03 \pm 1.90}$ & $\mathbf{48.18 \pm 1.81}$
		\\
		$\gamma$-PSO-LSTM & $\mathbf{48.99 \pm 1.84}$ & $\mathbf{48.95 \pm 1.73}$ & $\mathbf{48.97 \pm 1.71}$ & $\mathbf{48.13 \pm 1.77}$
		\\ \midrule \midrule
		LSTM & $\mathbf{\underline{55.31 \pm 2.64}}$ & $\mathbf{\underline{50.24 \pm 1.48}}$ & $\mathbf{\underline{48.89 \pm 1.71}}$ & $\mathbf{\underline{48.41 \pm 1.92}}$
		\\
		$\alpha$-GA-LSTM & $\mathbf{54.90 \pm 2.85}$ & $\mathbf{49.95 \pm 1.36}$ & $\mathbf{48.44 \pm 1.89}$ & $\mathbf{47.95 \pm 2.08}$
		\\
		$\alpha$-PSO-LSTM & $\mathbf{54.93 \pm 2.86}$ & $\mathbf{49.98 \pm 1.38}$ & $\mathbf{48.47 \pm 1.93}$ & $\mathbf{47.99 \pm 2.13}$
		\\
		$\beta$-GA-LSTM & $\mathbf{54.91 \pm 2.87}$ & $\mathbf{49.96 \pm 1.38}$ & $\mathbf{48.45 \pm 1.91}$ & $\mathbf{47.96 \pm 2.10}$
		\\
		$\beta$-PSO-LSTM & $\mathbf{54.92 \pm 2.89}$ & $\mathbf{49.97 \pm 1.36}$ & $\mathbf{48.45 \pm 1.93}$ & $\mathbf{47.98 \pm 2.13}$
		\\
		$\gamma$-GA-LSTM & $\mathbf{54.90 \pm 2.87}$ & $\mathbf{49.96 \pm 1.39}$ & $\mathbf{48.44 \pm 1.93}$ & $\mathbf{47.96 \pm 2.12}$
		\\
		$\gamma$-PSO-LSTM & $\mathbf{54.95 \pm 2.88}$ & $\mathbf{50.00 \pm 1.38}$ & $\mathbf{48.48 \pm 1.94}$ & $\mathbf{48.00 \pm 2.14}$
		\\ \bottomrule
	\end{tabular}
\end{table*}

Nonetheless, we think that meta-heuristics' performance was hindered by their high dimensional space (thousands of features), which often bottlenecks the exploration and exploitation phases as a low number of agents (roughly $10^2$ smaller order of magnitude) are not capable of traversing local optima and hence not converging to the most proper locations. On the other hand, the experimental results showed that meta-heuristics could sway the pre-trained weights in search of better values, which fosters feasible improvements and the necessity of additional experimentations.

\subsection{How Weights Influence the Fine-Tuning?}
\label{ss.weight}

%
%

\begin{figure*}
	\begin{tabular}{cc}
		\centering
		\includegraphics[scale=0.35]{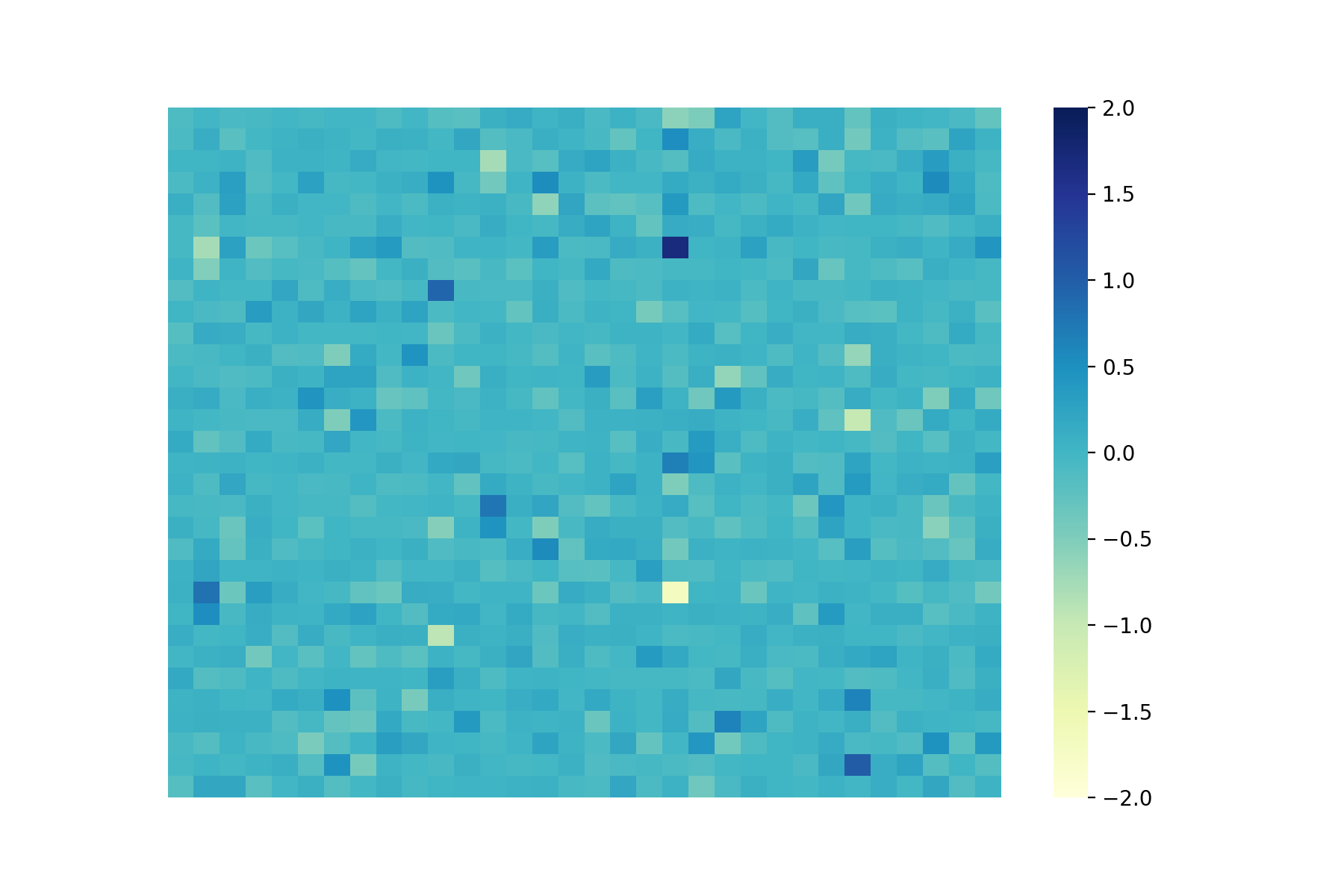} &
		\includegraphics[scale=0.35]{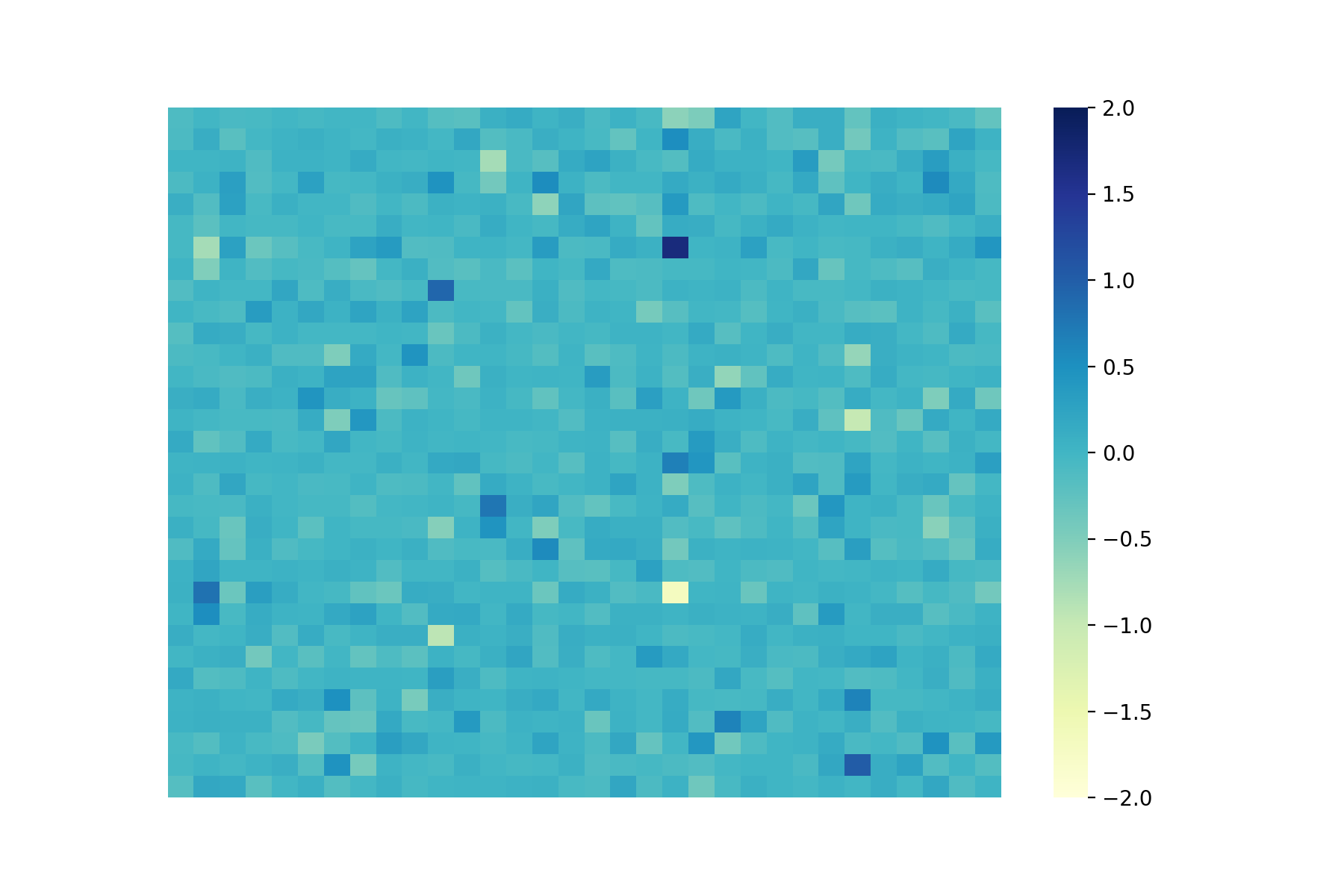}
		\\
		(a) & (b)
	\end{tabular}
	\caption{IMDB weights matrix comparison between: (a) LSTM and (b) $\gamma$-GA-LSTM.}
	\label{f.imdb_map}
\end{figure*}

\begin{figure*}
	\begin{tabular}{cc}
		\centering
		\includegraphics[scale=0.35]{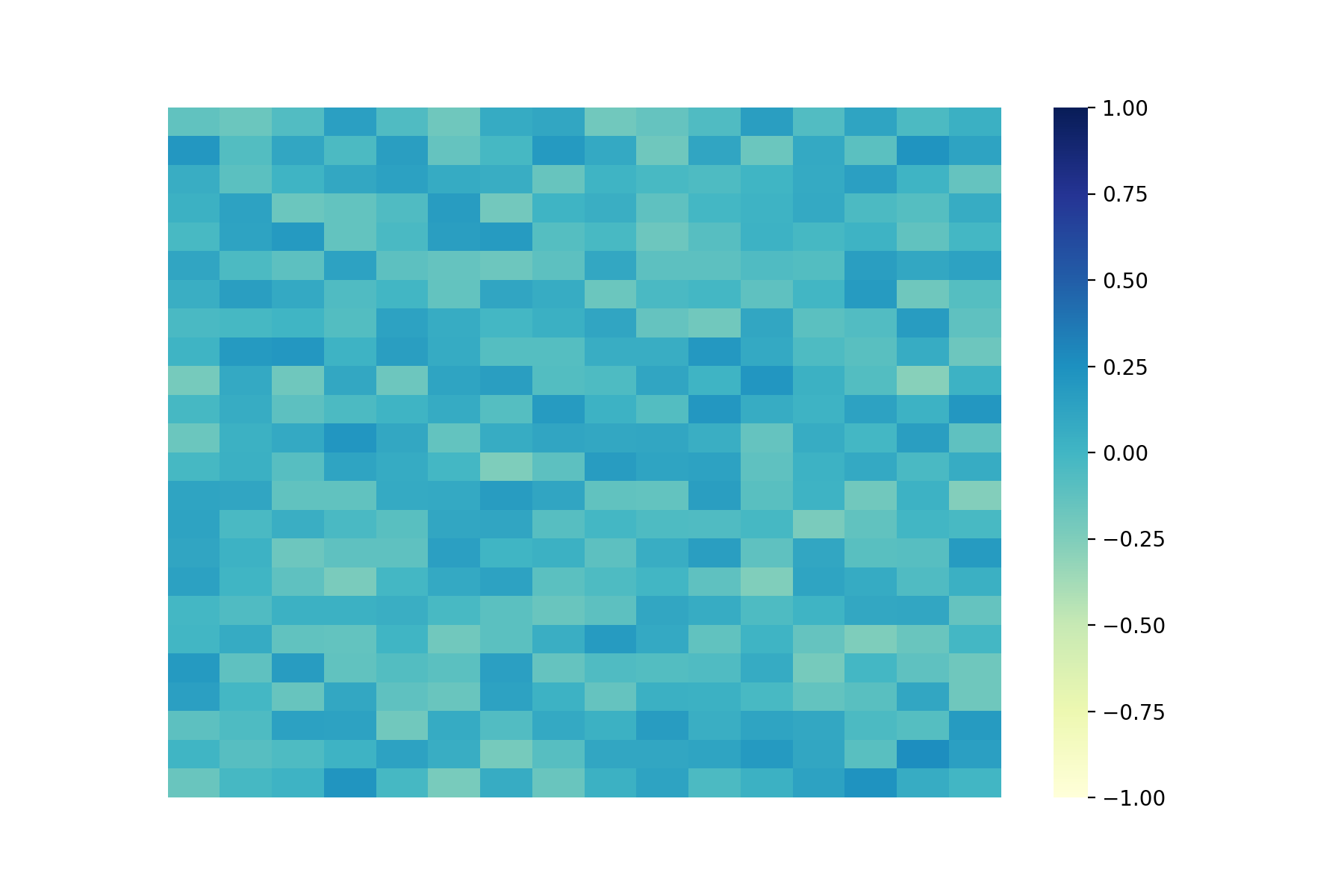} &
		\includegraphics[scale=0.35]{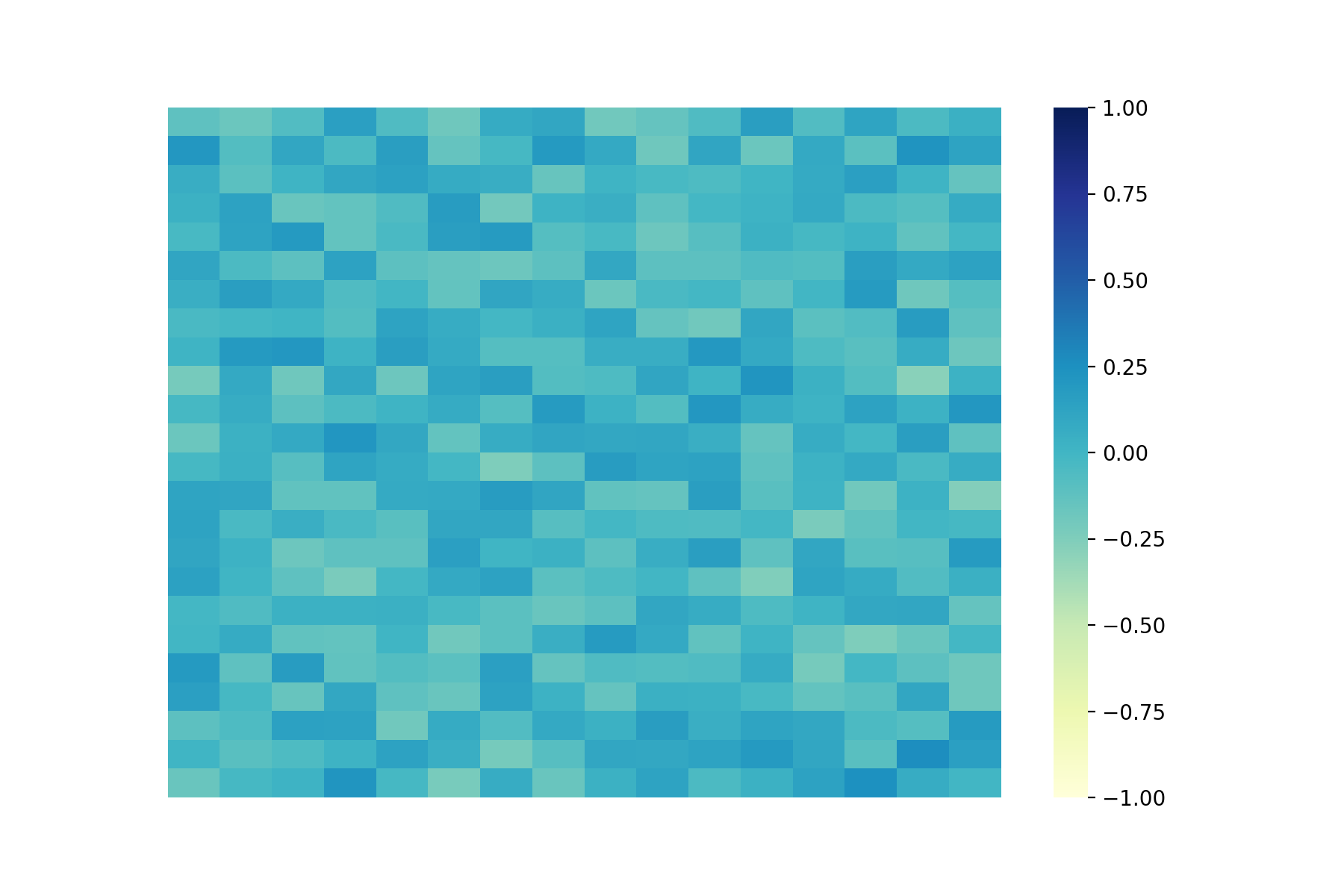}
		\\
		(a) & (b)
	\end{tabular}
	\caption{SST weights matrix comparison between: (a) LSTM and (b) $\gamma$-PSO-LSTM.}
	\label{f.sst_map}
\end{figure*}

Figure~\ref{f.imdb_map} illustrates the weights distribution concerning LSTM and $\gamma$-GA-LSTM models over the IMDB dataset, while Figure~\ref{f.sst_map} depicts the weights concerning LSTM and $\gamma$-PSO-LSTM models over the SST dataset. In both figures, it is hard to observe differences between plots (a) and (b), although plot (b) has slightly lighter colors when compared to plot (a). Such behavior indicates that the meta-heuristic algorithms found local optima when searching for weights values inferior to the pre-trained ones.

Furthermore, the absence of a critical difference between both plots indicates that the meta-heuristics could not adequately explore the search space, only attaining minimal superior or inferior values. As previously mentioned, this may happen due to an optimization in a not so ``important" layer, hence marginally improving or decreasing the optimized networks' performance.

\subsection{Bounding the Search Space}
\label{ss.search_space}

Finally, we opted to conduct the last experiment to verify the influence of distinct search bounds over the proposed methodology. Table~\ref{t.image_bounds} describes the accuracy results between distinct search bounds ($\Delta=0.0001$ and $\Delta=0.001$) over CIFAR-10 (top) and CIFAR-100 (bottom) datasets. Even though a larger $\Delta$ provided the best results (underlined cells), every model has been statistically similar to each other according to the Wilcoxon signed-rank test (bolded cells). Such performance strengthens the hypothesis that optimizing high dimensional search spaces requires more complex structures, such as increased agents and iterations, to accomplish an adequate convergence.

\begin{table}[!ht]
	\centering
	\setlength{\tabcolsep}{10pt}
	\renewcommand{\arraystretch}{1.75}
	\caption{Accuracy results ($\%$) between distinct search bounds ($\Delta$) over CIFAR-10 (top) and CIFAR-100 (bottom) datasets.}
	\label{t.image_bounds}
	\begin{tabular}{lcc}
		\toprule
		\textbf{Model} & $\mathbf{\Delta=0.0001}$ & $\mathbf{\Delta=0.001}$ 
		\\ \midrule
		$\alpha$-GA-MLP & $\mathbf{52.53 \pm 0.57}$ & $\mathbf{\underline{52.54 \pm 0.52}}$
		\\
		$\alpha$-PSO-MLP & $\mathbf{52.53 \pm 0.58}$ & $\mathbf{52.51 \pm 0.56}$
		\\
		$\beta$-GA-MLP & $\mathbf{52.53 \pm 0.58}$ & $\mathbf{52.49 \pm 0.57}$
		\\
		$\beta$-PSO-MLP & $\mathbf{52.52 \pm 0.56}$ & $\mathbf{52.53 \pm 0.56}$
		\\
		$\gamma$-GA-MLP & $\mathbf{\underline{52.54 \pm 0.57}}$ & $\mathbf{52.52 \pm 0.59}$
		\\
		$\gamma$-PSO-MLP & $\mathbf{52.51 \pm 0.58}$ & $\mathbf{52.52 \pm 0.59}$
		\\ \midrule \midrule
		$\alpha$-GA-MLP & $\mathbf{24.94 \pm 0.31}$ & $\mathbf{\underline{24.96 \pm 0.33}}$
		\\
		$\alpha$-PSO-MLP & $\mathbf{24.94 \pm 0.32}$ & $\mathbf{24.93 \pm 0.31}$
		\\
		$\beta$-GA-MLP & $\mathbf{24.93 \pm 0.31}$ & $\mathbf{24.91 \pm 0.33}$
		\\
		$\beta$-PSO-MLP & $\mathbf{24.92 \pm 0.31}$ & $\mathbf{\underline{24.96 \pm 0.34}}$
		\\
		$\gamma$-GA-MLP & $\mathbf{24.93 \pm 0.31}$ & $\mathbf{24.92 \pm 0.31}$
		\\
		$\gamma$-PSO-MLP & $\mathbf{24.94 \pm 0.32}$ & $\mathbf{24.92 \pm 0.31}$
		\\ \bottomrule
	\end{tabular}
\end{table}

Moreover, Table~\ref{t.text_bounds} describes the accuracy results between distinct search bounds ($\Delta=0.0001$ and $\Delta=0.001$) over IMDB (top) and SST (bottom) datasets. In such an experiment, it is possible to observe the same behavior depicted by Table~\ref{t.image_bounds}, where every meta-heuristic has been statistically similar to each other. Notwithstanding, one can also perceive that $\gamma$-based models with $\Delta=0.001$ achieved the best results concerning both datasets, supporting that LSTMs may benefit from broader than narrower searches.

\begin{table}[!ht]
	\centering
	\setlength{\tabcolsep}{10pt}
	\renewcommand{\arraystretch}{1.75}
	\caption{Accuracy results ($\%$) between distinct search bounds ($\Delta$) over IMDB (top) and SST (bottom) datasets.}
	\label{t.text_bounds}
	\begin{tabular}{lcccc}
		\toprule
		\textbf{Model} & $\mathbf{\Delta=0.0001}$ & $\mathbf{\Delta=0.001}$
		\\ \midrule
		$\alpha$-GA-LSTM & $\mathbf{48.98 \pm 1.69}$ & $\mathbf{48.98 \pm 1.76}$
		\\
		$\alpha$-PSO-LSTM & $\mathbf{48.98 \pm 1.72}$ & $\mathbf{48.97 \pm 1.54}$
		\\
		$\beta$-GA-LSTM & $\mathbf{48.98 \pm 1.85}$ & $\mathbf{48.98 \pm 1.93}$
		\\
		$\beta$-PSO-LSTM & $\mathbf{48.98 \pm 1.73}$ & $\mathbf{48.98 \pm 1.88}$
		\\
		$\gamma$-GA-LSTM & $\mathbf{48.98 \pm 1.77}$ & $\mathbf{\underline{49.03 \pm 1.88}}$
		\\
		$\gamma$-PSO-LSTM & $\mathbf{48.97 \pm 1.69}$ & $\mathbf{48.99 \pm 1.84}$
		\\ \midrule \midrule
		$\alpha$-GA-LSTM & $\mathbf{54.91 \pm 2.88}$ & $\mathbf{54.90 \pm 2.85}$
		\\
		$\alpha$-PSO-LSTM & $\mathbf{54.91 \pm 2.88}$ & $\mathbf{54.93 \pm 2.86}$
		\\
		$\beta$-GA-LSTM & $\mathbf{54.91 \pm 2.88}$ & $\mathbf{54.91 \pm 2.87}$
		\\
		$\beta$-PSO-LSTM & $\mathbf{54.91 \pm 2.88}$ & $\mathbf{54.92 \pm 2.89}$
		\\
		$\gamma$-GA-LSTM & $\mathbf{54.91 \pm 2.88}$ & $\mathbf{54.90 \pm 2.87}$
		\\
		$\gamma$-PSO-LSTM & $\mathbf{54.91 \pm 2.88}$ & $\mathbf{\underline{54.95 \pm 2.88}}$
		\\ \bottomrule
	\end{tabular}
\end{table}

%% file: sections/conclusion.tex
\section{Conclusion}
\label{s.conclusion}

This work presented an early draft of how to fine-tune neural networks' performance through meta-heuristic techniques. Essentially, after training architectures through standard gradient descent algorithms, meta-heuristic algorithms attempt to explore the trained search space and find more suitable positions.

The experimental results showed that Multi-Layer Perceptron applied to image classification tasks could benefit from weights fine-tuning. While the meta-heuristics (GA and PSO) performance were statistically similar to the standard trained architecture, they could attain a higher mean accuracy and a lower standard deviation over the CIFAR-10 dataset ($52.54 \pm 0.52$ against $52.40 \pm 0.62$) while obtaining a higher mean accuracy and higher standard deviation over the CIFAR-100 dataset ($24.96 \pm 0.33$ against $24.93 \pm 0.30$). Such results strengthen the ability of meta-heuristics to fine-tune and better adjust the learned weights.

Regarding the sentiment analysis task conducted by the LSTMs, one can perceive that none meta-heuristic could obtain better metrics than the standard architecture, yet all results were statistically similar according to the Wilcoxon signed-rank test. Such behavior might be explained due to only fine-tuning the last fully-connected layer (layer before the Softmax activation) instead of fine-tuning the recurrent layer and ignoring the biases fine-tuning, which may help meta-heuristics in achieving more competitive results.

Finally, glimpsing through future works, we aim to extend the current framework by using new sets of meta-heuristics algorithms, including fine-tuned biases, and extending the number of fine-tuned layers. Additionally, we aim at extending the current work to other neural network architectures, such as Convolutional Neural Networks and Transformers. We believe that the key to proper fine-tuning lies in selecting the most informative layer (a layer that is responsible for extracting the most important information) and a not so overwhelming search space (fewest feature spaces).